# Feasibility of Principal Component Analysis in hand gesture recognition system

Tanu Srivastava, Raj Shree Singh, Sunil Kumar
Dept. of Computer Science and Engineering
Mody University of Science and Technology
Lakshmangarh, Rajasthan, India
tanusri.ts@gmail.com,rjshree91@gmail.com,skvasistha@gmail.com

Pavan Chakraborty
Assistant Professor (Dept. of Robotics and AI)
Indian Institute of Information & Technology
Allahabad, India
pavan.chakraborty@gmail.com

**ABSTRACT**

Nowadays actions are increasingly being handled in electronic ways, instead of physical interaction. From earlier times biometrics is used in the authentication of a person. It recognizes a person by using a human trait associated with it like eyes (by calculating the distance between the eyes) and using hand gestures, fingerprint detection, face detection etc. Advantages of using these traits for identification are that they uniquely identify a person and cannot be forgotten or lost. These are unique features of a human being which are being used widely to make the human life simpler. Hand gesture recognition system is a powerful tool that supports efficient interaction between the user and the computer. The main moto of hand gesture recognition research is to create a system which can recognise specific hand gestures and use them to convey useful information for device control. This paper presents an experimental study over the feasibility of principal component analysis in hand gesture recognition system. PCA is a powerful tool for analyzing data. The primary goal of PCA is dimensionality reduction. Frames are extracted from the Sheffield KInect Gesture (SKIG) dataset. The implementation is done by creating a training set and then training the recognizer. It uses Eigen space by processing the eigenvalues and eigenvectors of the images in training set. Euclidean distance with the threshold value is used as similarity metric to recognize the gestures. The experimental results show that PCA is feasible to be used for hand gesture recognition system.

## 1. INTRODUCTION

Gestures originating from face and hand are most commonly used in the applications that work in Human Gesture Recognition technology. These techniques have already established hand as complex and important structure to be used by recognition technology. But now a days hand gestures also have gained interests of different device manufacturers apart from their application in Telerobotics, sign language recognition and Human Computer Interaction (HCI). Face recognition systems although very popular has some disadvantages.

(i) Problem with false rejection during recognition when people change their hair style, grow or shave a beard or wear glasses.

(ii) It can't tell the difference between identical twins.

There are two types of gestures namely static and dynamic. Static gesture means user with certain fix pose or gesture. Dynamic means gesture with some strokes they can be either prestroke, stroke or poststroke phases. For static gestures, images are processed directly after converting them to vector form but for videos or dynamic gestures the frames are first extracted from videos then processed after converting them to vector form.

In this paper the focus is on dynamic gestures. The frames are extracted from Sheffield Kinect Gesture (SKIG) dataset[6]. This dataset collects 10 categories of hand gestures i.e. circle, triangle, up-down, right-left wave, "Z", cross and many more. In this process all ten categories are performed with three hand gestures; fist, index and flat, and recorded the sequence under three different backgrounds i.e. wooden board, white plain paper and paper with characters.

The algorithm used is Principal Component Analysis, which is a powerful tool for analysing data and its main feature is dimensionality reduction. Dimensionality reduction by PCA helps in the following two ways;

(i) It avoids a lot of huge calculations.

(ii) It is robust to the noise in representation of images.

If dimensionality reduction is not performed the size of calculated eigenvectors would be so large that it would terribly slow down the system or run out of memory as computation required would be huge. PCA uses the Eigen space by calculating the eigenvalues and eigenvectors of the gesture images in the training set. The eigenvectors are calculated from the covariance matrix after performing dimensionality reduction on covariance matrix, usually after normalizing (mean centring) the data matrix for each attribute. PCA results discussed in terms of principal components, and respective weights.

This paper is organized as follows; Section 2 summarizes literature related to different hand gesture recognition systems. Section 3 describes procedure which includes creating a training set and loading it, training the recognizer, reducing the dimensionality of training set, processing eigenvectors and eigenvalues and selecting k-significant eigenvectors and then final recognition process by comparing the Euclidean distance with some threshold value. Section 4 includes results, Section 5 conclude the paper.

## 2. LITERATURE SURVEY

Many forms of work have been done using PCA and Eigen image approach. Different line of action for data collection has been given nicely by Rama B. Dan, P. S. Mohod in [1] and each approach has situation based advantage. Recently, growing interest has been shown by researchers in it. As discussed earlier about the types of gestures, dynamic gestures are more commonly analyzed and needed as a medium of Human Computer Interaction (HCI), but many a times static gesture recognition forms the foundation for the former. Mule, K. C., & Holambe in [2] enlists detailed information on hand gesture recognition using two methods, firstly, using Histogram projection for feature extraction and secondly using PCA. The paper, however, concluded that a combination of PCA and projection yields more accurate results than applying PCA and projection separately. Shitole, S. M., Patil, S. B., & Narote in [3] satisfactorily describes the steps followed in dynamic gesture recognition with 2 different systems. First involves RGB based color segmentation followed by appearance based modelling to extract the features. It also states the easiness of getting templates after segmentation and classification of these segments using PCA. However, it has certain disadvantages in the form of restriction to user and poor performance. Second system was developed without colored markers but used skin color detection and discussions showed that ANN provided much better classification. The results were also found more accurate than the previous system. Agrawal T. & Chaudhuri S. in [4] used PCA of feature trajectory to recognize unknown gestures approach and had a significant advantage in terms of rate at which gestures were performed. It was computationally feasible as PCA aids in dimensionality reduction of feature space and good accuracy was obtained with small number of principal components. Jeong, D. H. and Ziemkiewicz, C. and Ribarsky, W. and Chang in [5] provides a detailed description of PCA based on a designed visual analytics tool (DVA) that visualizes the results of PCA and supports a set of interactions to help the user in better understanding and utilizing PCA.

## 3. METHOD

### 3.1 TRAINING SET PREPARATION

The training set consist of 10 categories of hand gestures in total; circle (clockwise), triangle (anti-clockwise), up-down, right-left, wave, "Z", cross, come-here, turn-around and pat. All these gestures are performed with three hand postures: fist, index and flat under three different backgrounds (i.e., wooden board, plain paper and paper with characters). These gestures are extracted in frames from SKIG dataset.

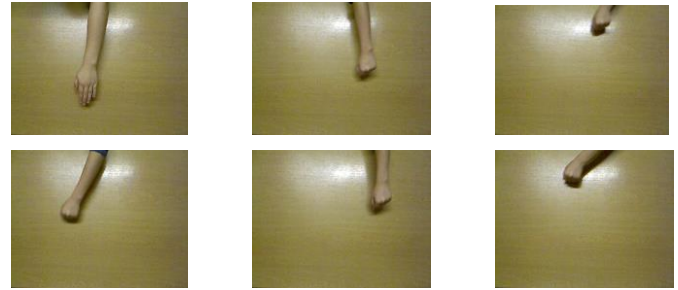

**Fig. 3.1 Gestures with wooden background**

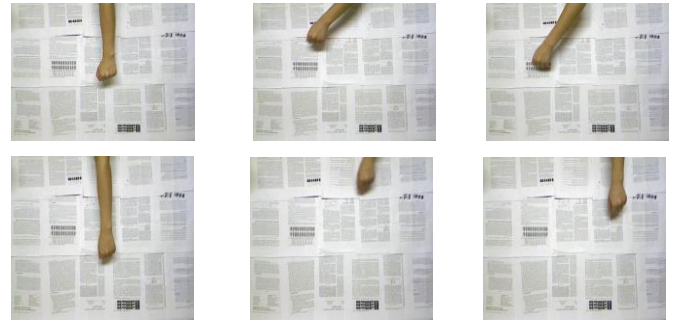

**Fig. 3.2 Gestures with characters in background**

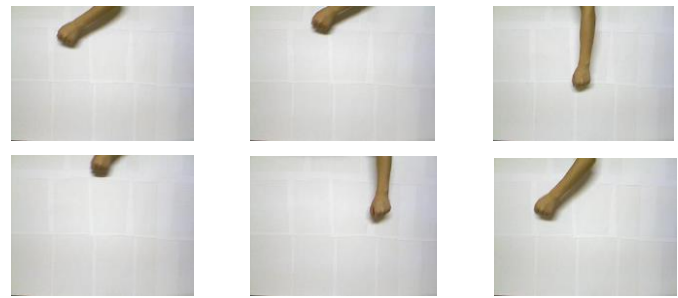

**Fig. 3.3 Gestures with plain paper background**

### 3.2 ALGORITHM

The procedure of hand gesture recognition using principal component analysis involves the following steps;

Step I: Creating training set and loads it

There are total m images in the training set each of size n*n.

Step II: Train the recognizer

(i) Convert gesture images in the training set to vector form, as principal component analysis does not work on the images directly but convert it into a vector form. Images are converted into a column vector.

(ii) Normalise the gesture image vectors

Normalisation removes all the common features that the gesture images share together so that every image vector left behind with unique features. Calculate the average image vector i.e. φ. Subtract average from each gesture image vector.

Normalized face vector,

$$\varphi_i = I_i - \varphi \qquad \ldots\ldots\text{Eq. 3.1}$$

where $\varphi_i$ = normalized face vector

$I_i$ = gesture image vector

$\Phi$ = average image vector

(iii) Calculate the Eigen vector by calculating the covariance matrix C

$$C = A * A^T \qquad \ldots\ldots\text{Eq. 3.2}$$

where A $[\varphi_1, \varphi_2, \varphi_3 \ldots \varphi_M]$, matrix column with each normalized gesture.

So, $A = N^2 * M \qquad \ldots\ldots\text{Eq. 3.3}$

where $N^2$ is number of rows and M is the number of columns.

Therefore dimension of C becomes $N^2 * N^2$, which will generate $N^2$ eigenvectors which is as big as the gesture image itself.

(iv) The need for dimensionality reduction

The dimension of Covariance matrix is very large hence generating large sized eigenvectors. This may terribly slow down the system or run out of memory as the computations required are huge.

To reduce the calculation and effect of noise on the needed eigenvectors, calculate them from a covariance matrix of reduced dimensionality.

Now, covariance matrix would be calculated as:

$$C = A^T * A \qquad \ldots\ldots\text{Eq. 3.4}$$

Hence giving the dimension of C as M*M.

(v) Calculate eigenvectors from covariance matrix

Total number of rows in Covariance matrix gives the total number of eigenvectors.

Hence total number of eigenvectors = M, each of M×1 dimensionality where $M << N^2$.

(vi) K-best Eigen images are selected such that K<M and can represent the whole training set.

After selecting the Eigen images, they are mapped back to original dimensionality from "M×1" to "$N^2 \times N^2$".

$$U_i = A * V_i \qquad \ldots\ldots\text{Eq. 3.5}$$

where $U_i$ – the Eigen vector in higher dimensional space

$V_i$ – the Eigen vector in lower dimensional space

(vii) Represent each gesture image as a linear combination of all K-eigenvectors.

For each gesture image in training set we will represent the image as a linear combination of these K-Eigen images plus the mean image (average image) as earlier the gesture image vector was subtracted from the average gesture image vector. So we have to add back the average gesture image to get all the features back.

$\Omega$ – a weight vector, calculated for each gesture image

And for each image in the training set, associated weight vectors are calculated and stored.

(viii) Calculation of Euclidean distance

The Euclidean distance between input weight vector and all the weight vectors of training set are calculated and compared with a threshold distance. If the calculated distance is less than the threshold value then the gesture is known otherwise unknown.

$$\Omega = \begin{pmatrix} w_1 \\ w_2 \\ w_3 \\ w_k \end{pmatrix}$$

where $w_1, w_2, w_3, \ldots, w_k$ are the associated weights with each input gesture vector

The complete procedure adopted in recognition process is shown in the following block diagram:

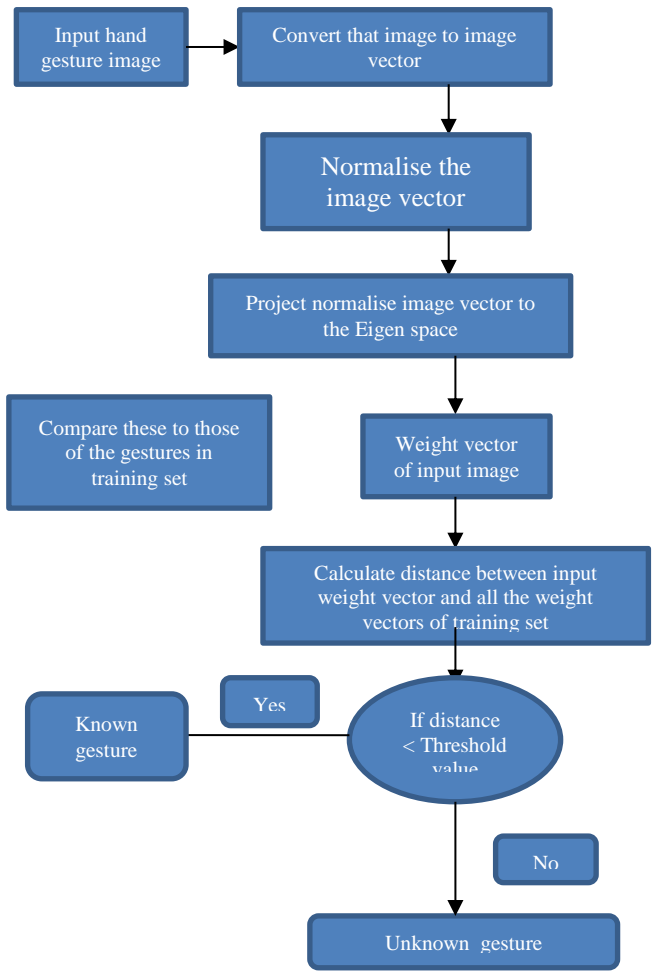

Fig. 3.4 Recognition process framework

## 4. RESULTS

After the recognition process the system successfully calculates the efficiency by finding out the total number of hand gestures matched.

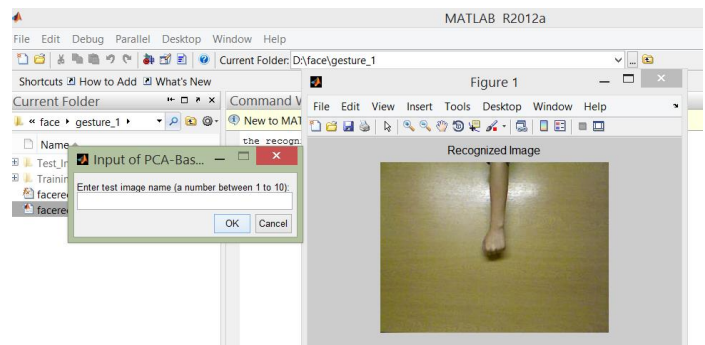

Fig. 4.1 Inputting test images

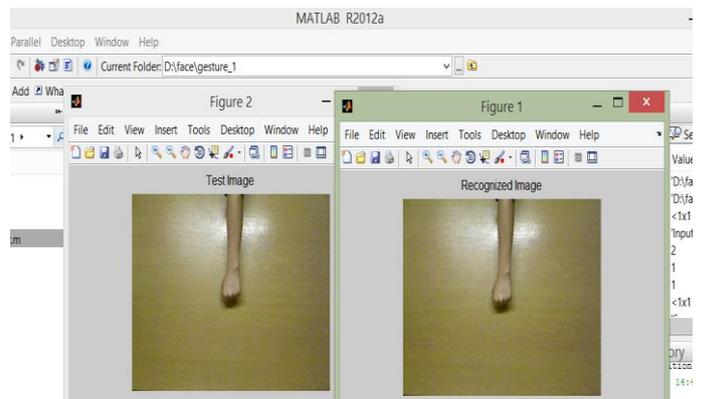

Fig. 4.2 Matched images

The calculated efficiency is shown in the following snapshot:

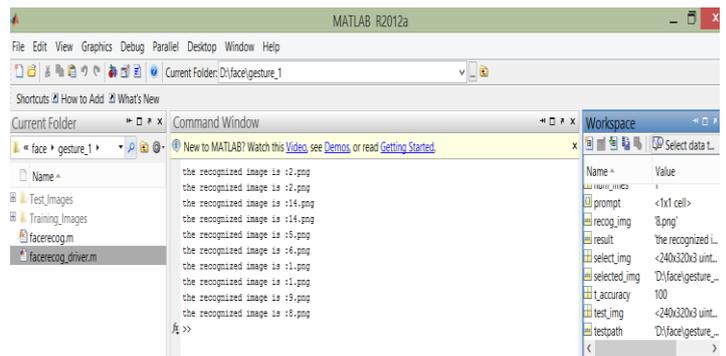

Fig.4.3 The snapshot depicts the total calculated accuracy

Table 4.1 Accuracy test on controlled background images

| Condition of Image | Total Tested | Successful Detection | Failures | Accuracy |
|---|---|---|---|---|
| Controlled background | 10 | 10 | 0 | 100% |

Table 4.2 Statistical analysis of proposed model results

| PROPERTY | VALUE |
|---|---|
| True Positive | 10 |
| True Negative | 0 |
| False Positive | 0 |
| False Negative | 0 |
| Sensitivity (Recall) | 100% |
| Precision (Positive Predictive) | 100% |
| Prevalence | 100% |

**Formulas for calculating Recall, Precision and Prevalence**

$$\text{Recall} = \frac{\text{True Positive}}{(\text{True Positive} + \text{False Negative})}$$

$$\text{Precision} = \frac{\text{True Positive}}{(\text{True Positive} + \text{False Positive})}$$

$$\text{Prevalence} = \frac{\text{(True Positive + False Negative)}}{\text{Total Number of Gestures}}$$

Table 4.1 and 4.2 show that accuracy, sensitivity prevalence and Precision achieved in result are 100%. This is due to the fact that the test dataset and training dataset are almost similar.

## 5. CONCLUSION

In this paper recognition problem is solved through a matching process in which a test set hand image is compared with all the images in database. A novel approach of hand gesture recognition is proposed by using PCA for feature extraction. PCA is acknowledged as a robust method which handles images even having noise (variation of brightness or color information) and more compression ratio (as opposed to pixel to pixel matching process in which no noise could be handled). The principal components itself signifies the most prevalent features in the test set. Thus a greater efficiency is achieved in each test of images. However, the model proposed here does not evaluate images with uncontrolled background. Also the model works only with static gestures. In future the system can be upgraded to work for uncontrolled background and the efficiency of PCA can be accentuated.